\documentclass[10pt,twocolumn,letterpaper]{article}

\usepackage{iccv}
\usepackage{times}
\usepackage{epsfig}
\usepackage{graphicx}
\usepackage{amsmath}
\usepackage{amssymb}
\usepackage{caption}
\usepackage{subcaption}

\usepackage{booktabs}
\usepackage{multirow}
\usepackage{array}
\usepackage{float}
\usepackage{bbm}

\newcommand{\drule}{\specialrule{0.2pt}{1pt}{1pt}%
            \specialrule{0.2pt}{0pt}{\belowrulesep}%
            }
  {\begin{list}{}%
          {\setlength{\leftmargin}{#1}}%
          \item[]%
  }
  {\end{list}}

\usepackage{pifont}
\usepackage{xcolor}
\usepackage{pdfrender}
\newcommand{\xmark}{\ding{55}}%
\usepackage{colortbl}
  
\newcommand*{\affmark}[1][*]{\textsuperscript{#1}}

\usepackage{kotex}

\definecolor{applegreen}{rgb}{0.55, 0.71, 0.0}
\newcommand*{\boldcheckmark}{%
  \textpdfrender{
    TextRenderingMode=FillStroke,
    LineWidth=.5pt, 
  }{\checkmark}%
}
\newcommand*{\boldxmark}{%
  \textpdfrender{
    TextRenderingMode=FillStroke,
    LineWidth=.5pt, 
  }{\xmark}%
}
\definecolor{forestgreen}{rgb}{0.13, 0.55, 0.13}
\DeclareMathOperator*{\argmax}{arg\,max}

\newcommand\blfootnote[1]{%
  \begingroup
  \renewcommand\thefootnote{}\footnote{#1}%
  \addtocounter{footnote}{-1}%
  \endgroup
}

\usepackage[breaklinks=true,colorlinks,bookmarks=false]{hyperref}

\iccvfinalcopy 

\def\iccvPaperID{2171} 

\ificcvfinal\fi
\begin{document}

\title{Standardized Max Logits: A Simple yet Effective Approach for Identifying Unexpected Road Obstacles in Urban-Scene Segmentation}


\author{
Sanghun Jung$^\text{*}$\affmark[1] \; Jungsoo Lee$^\text{*}$\affmark[1] \; Daehoon Gwak\affmark[1] \; Sungha Choi\affmark[2] \; Jaegul Choo\affmark[1]\vspace{0.2cm}\\
\affmark[1]KAIST AI \; \affmark[2]LG AI Research\vspace{0.2cm}\\
\texttt{\footnotesize\affmark[1]\{shjung13, bebeto, daehoon.gwak, jchoo\}@kaist.ac.kr} \; \texttt{\footnotesize\affmark[2]shachoi@korea.ac.kr}\\
\vspace*{-1.5cm}
}

\makeatletter
\g@addto@macro\@maketitle{
  \begin{figure}[H]
  \setlength{\linewidth}{\textwidth}
  \setlength{\hsize}{\textwidth}
  \centering
  \includegraphics[width=1.0\linewidth]{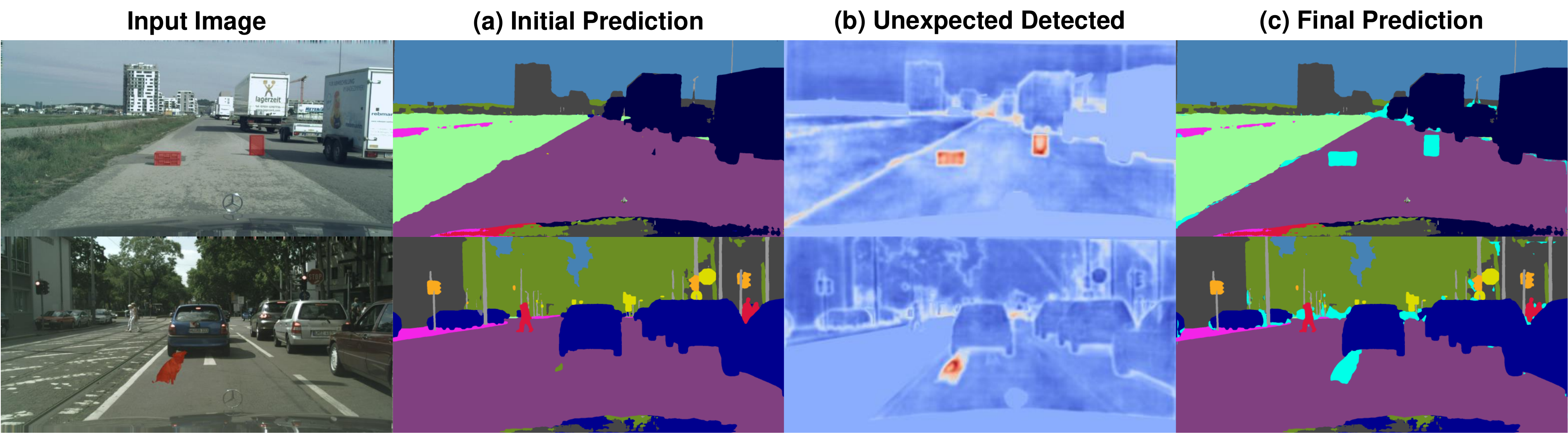}
   \vspace*{-0.7cm}
    \captionof{figure}{Results of our approach on identifying unexpected obstacles on roads. (a) Previous segmentation networks classify the unexpected obstacles (\textit{e.g.,} dogs) as one of the pre-defined classes (\textit{e.g.,} road) which may be detrimental from the safety-critical perspective. (b) Through our method, we detect the unexpected obstacles. (c) Finally, we can obtain the final prediction of segmentation labels with unexpected obstacles (cyan-colored objects) identified.}
  \label{fig:main_figure}
  \vspace*{-0.3cm}
  \end{figure}
}
\makeatother

\maketitle
\blfootnote{\vspace*{-0.8cm}* indicates equal contribution}

\ificcvfinal\thispagestyle{empty}\fi

\vspace{-0.2cm}
\begin{abstract}
\vspace{-0.3cm} 
Identifying unexpected objects on roads in semantic segmentation (\textit{e.g.,} identifying dogs on roads) is crucial in safety-critical applications. 
Existing approaches use images of unexpected objects from external datasets or require additional training (e.g., retraining segmentation networks or training an extra network), which necessitate a non-trivial amount of labor intensity or lengthy inference time.
One possible alternative is to use prediction scores of a pre-trained network such as the max logits (i.e., maximum values among classes before the final softmax layer) for detecting such objects.
However, the distribution of max logits of each predicted class is significantly different from each other, which degrades the performance of identifying unexpected objects in urban-scene segmentation.
To address this issue, we propose a simple yet effective approach that \textbf{standardizes} the max logits in order to align the different distributions and reflect the relative meanings of max logits within each predicted class.  
Moreover, we consider the local regions from two different perspectives based on the intuition that neighboring pixels share similar semantic information. 
In contrast to previous approaches, our method does not utilize any external datasets or require additional training, which makes our method widely applicable to existing pre-trained segmentation models. Such a straightforward approach achieves a new state-of-the-art performance on the publicly available Fishyscapes Lost \& Found leaderboard with a large margin. Our code is publicly available at this link\footnote{\vspace*{-0.8cm}https://github.com/shjung13/Standardized-max-logits}.
\end{abstract}
\vspace{-0.4cm}

\vspace{-0.1cm}
\section{Introduction}
\vspace{-0.1cm}

\begin{figure*}[ht!]
  \includegraphics[width=\linewidth]{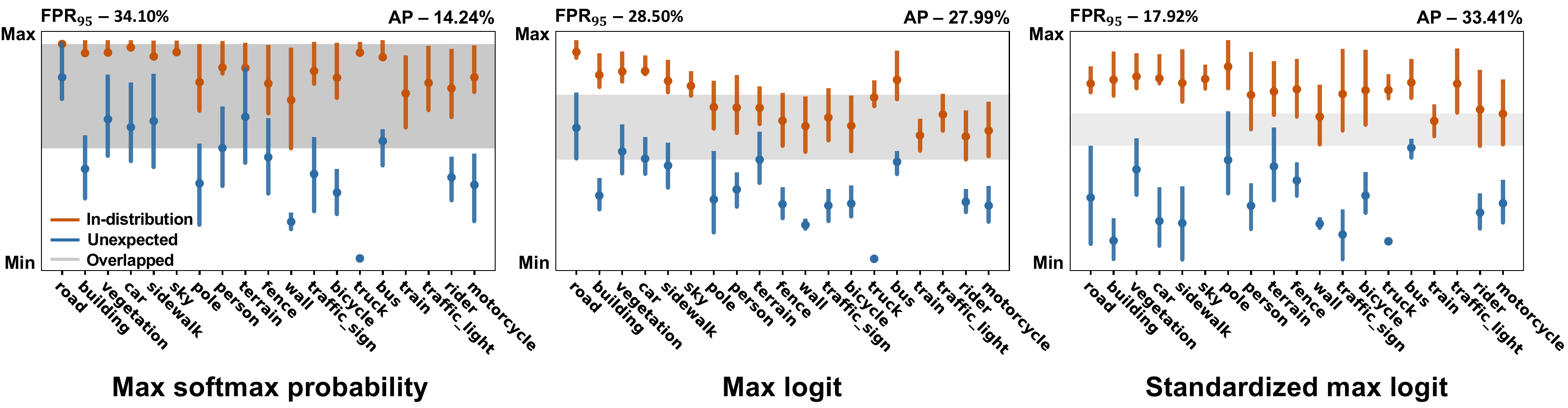}
  \vspace*{-0.5cm}
    \caption{Box plots of MSP, max logit, and standardized max logit in Fishyscapes Static. X-axis denotes the classes which are sorted by the occurrences of pixels in the training phase. Y-axis denotes the values of each method. \textcolor{black}{Red and blue represent the distributions of values in in-distribution pixels and unexpected pixels, respectively. The lower and upper limits of each bar indicate the Q1 and Q3 while the dot represents the mean value of its predicted class. The gray indicates the overlapped regions of the two groups}. The opacity of the gray region is proportional to the FPR at TPR 95\%. \textcolor{black}{Standardizing the max logits in a class-wise manner clearly reduces the FPR.}
    }
    \vspace*{-0.6cm}
    \label{fig:analysis_box_plot}
\end{figure*}

Recent studies~\cite{robustnet, hanet, foveanet, denseASPP, class_uniform_and_urban_scene, anlnet_urban_scene, danet_urban_scene} in semantic segmentation focus on improving the segmentation performance on urban-scene images. 
Despite such recent advances, these approaches cannot identify \emph{unexpected objects} (\textit{i.e.,} objects not included in the pre-defined classes during training), mainly because they predict all the pixels as one of the pre-defined classes.
Addressing such an issue is critical especially for safety-critical applications such as autonomous driving. As shown in Fig.~\ref{fig:main_figure}, wrongly predicting a dog (\textit{i.e.,} an unexpected object) on the road as the road does not stop the autonomous vehicle, which may lead to roadkill.
In this safety-critical point of view, the dog should be detected as an unexpected object which works as the starting point of the autonomous vehicle to handle these objects differently (\textit{e.g.,} whether to stop the car or circumvent the dog).

Several studies~\cite{fishyscapes, resynthesis, erasing, entropy, lost_and_found, dense, real-nvp} tackle the problem of detecting such unexpected objects on roads.  
Some approaches~\cite{dense, entropy} utilize external datasets~\cite{ILSVRC, COCO} as samples of unexpected objects while others~\cite{resynthesis, synthesize_compare, erasing, accv_road_obstacle} leverage image resynthesis models for erasing the regions of such objects.
However, such approaches require a considerable amount of labor intensity or necessitate a lengthy inference time. 
On the other hand, simple approaches which leverage only a pre-trained model~\cite{baseline, odin, mahalanobis} are proposed for out-of-distribution (OoD) detection in image classification, the task of detecting images from a different distribution compared to that of the train set.
Based on the intuition that a correctly classified image generally has a higher maximum softmax probability (MSP) than an OoD image~\cite{baseline}, MSP is used as the anomaly score \textcolor{black}{(\textit{i.e.,} the value used for detecting OoD samples).}
Alternatively, utilizing the max logit~\cite{maxlogit} (\textit{i.e.,} maximum values among classes before the final softmax layer) as the anomaly score is proposed, which outperforms using MSP for detecting anomalous objects in semantic segmentation. 
Note that \emph{high} prediction scores (\textit{e.g.,} MSP and max logit) indicate \emph{low} anomaly scores and vice versa. 
\vspace{-0.1cm}

However, directly using the MSP~\cite{baseline} or the max logit~\cite{maxlogit} as the anomaly score has the following limitations. 
Regarding the MSP~\cite{baseline}, the softmax function has the fast-growing exponential property which produces highly confident predictions.
\textcolor{black}{Pre-trained networks may be highly confident with OoD samples which limits the performance of using MSPs for detecting the anomalous samples~\cite{odin}.}  
In the case of the max logit~\cite{maxlogit}, as shown in Fig.~\ref{fig:analysis_box_plot}, the values of the max logit have their own ranges in each predicted class.
Due to this fact, the max logits of the unexpected objects predicted as particular classes (\textit{e.g.,} road) exceed those of other classes (\textit{e.g.,} train) in the in-distribution objects.
This can degrade the \textcolor{black}{performance} of detecting unexpected objects on evaluation metrics (\textit{e.g.,} AUROC and AUPRC) that use the same threshold for all classes.

In this work, inspired by this finding, we propose standardizing the max logits in a class-wise manner, termed \emph{standardized max logits} (SML). 
\textcolor{black}{Standardizing the max logits aligns the distributions of max logits in each predicted class, so it enables to reflect the relative meanings of values within a class. 
This reduces the false positives (\textit{i.e.,} in-distribution objects detected as the unexpected objects, highlighted as gray regions in Fig.~\ref{fig:analysis_box_plot}) when using a single threshold.}

Moreover, we further improve the performance of identifying unexpected obstacles using the local semantics from two different perspectives.
\textcolor{black}{First, we remove the false positives in boundary regions where predicted class changes from one to another.
Due to the class changes, the boundary pixels tend to have low prediction scores (\textit{i.e.,} high anomaly scores) compared to the non-boundary pixels~\cite{boundary_active, boundary_neural}.}
\textcolor{black}{In this regard}, we propose a novel \emph{iterative boundary suppression} to remove such false positives by replacing the high anomaly scores of boundary regions with low anomaly scores of neighboring \textcolor{black}{non-boundary pixels}.
\textcolor{black}{Second, in order to remove the remaining false positives in both boundary and non-boundary regions, we smooth them using the neighboring pixels based on the intuition that local consistency exists among the pixels in a local region. We term this process as \emph{dilated smoothing.}}
\begin{figure*}[ht!]
\begin{center}
    \includegraphics[width=\linewidth]{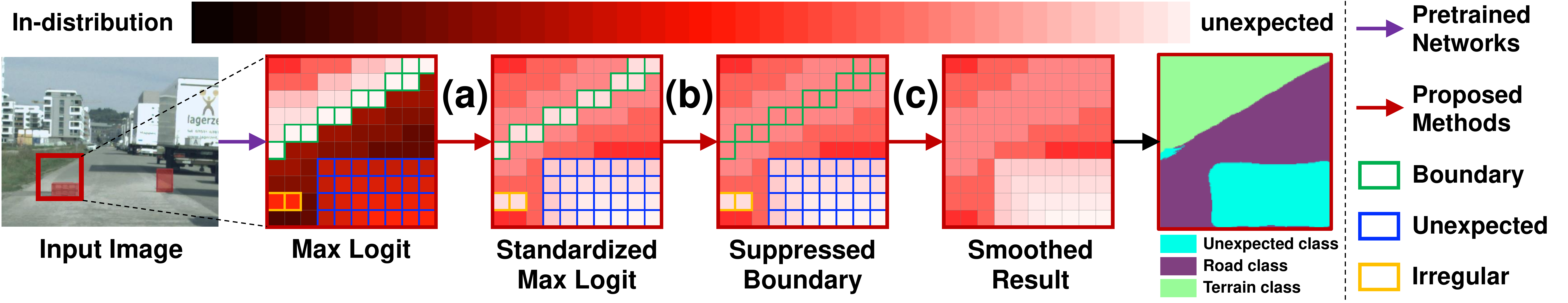}
\end{center}
\vspace*{-0.7cm}
\caption{Overview of our method. We obtain the max logits from a segmentation network and (a) standardize it using the statistics obtained from the training samples. (b) Then, we iteratively replace the standardized max logits of boundary regions with those of surrounding non-boundary pixels. (c) Finally, we apply dilated smoothing to consider local semantics in broad receptive fields.}
\label{fig:method_overview}
\vspace*{-0.6cm}
\end{figure*}

The main contributions of our work are as follows:
\begin{itemize}
    \vspace{-0.1cm}
    \item We propose a \textit{simple yet effective} approach for identifying unexpected objects on roads in urban-scene semantic segmentation.
    \vspace{-0.1cm}
    \item Our proposed approach can easily be applied to various existing models since our method does not require additional training or external datasets.
    \vspace{-0.1cm}
    \item We achieve a new state-of-the-art performance on the publicly available Fishyscapes Lost \& Found Leaderboard\footnote{\vspace{-0.8cm}https://fishyscapes.com/} among the previous approaches with a large margin and negligible computation overhead while not requiring additional training and OoD data.

\end{itemize}


\vspace{-0.35cm}
\section{Related Work}
\vspace{-0.1cm}
\subsection{Semantic segmentation on urban driving scenes}
\vspace{-0.1cm}
Recent studies~\cite{robustnet, hanet, foveanet, denseASPP, class_uniform_and_urban_scene, anlnet_urban_scene, danet_urban_scene, hardnet, efficient_fusion, bidirectional} have strived to enhance the semantic segmentation performance on urban scenes.
The studies~\cite{foveanet, denseASPP} consider diverse scale changes in urban scenes or leverage the innate geometry and positional patterns found in urban-scene images~\cite{hanet}. 
Moreover, several studies~\cite{hardnet, efficient_fusion, bidirectional} have proposed more efficient architectures to improve the inference time, which is critical for autonomous driving.
Despite the advances, unexpected objects cannot be identified by these models, which is another important task for safety-critical applications.
Regarding the importance of such a task from the safety-critical perspective, we focus on detecting unexpected obstacles in urban-scene segmentation.

\vspace{-0.1cm}
\subsection{Detecting unexpected objects in semantic segmentation}
\vspace{-0.1cm}
Several studies~\cite{dense, entropy, fishyscapes} utilize samples of unexpected objects from external datasets during the training phase.
For example, by assuming that the objects cropped from the ImageNet dataset~\cite{ILSVRC} are anomalous objects, they are overlaid on original training images~\cite{dense} (\textit{e.g.}, Cityscapes) to provide samples of unexpected objects. 
\textcolor{black}{Similarly, another previous work~\cite{entropy} utilizes the objects from the COCO dataset~\cite{COCO} as samples of unexpected objects.}
However, such methods require retraining the network by using the additional datasets, which hampers to utilize a given pre-trained segmentation network directly.

Other work~\cite{resynthesis, synthesize_compare, erasing, accv_road_obstacle} exploits the image resynthesis (\textit{i.e.,} reconstructing images from segmentation predictions) for detecting unexpected objects.
Based on the intuition that image resynthesis models fail to reconstruct the regions with unexpected objects, these studies use the discrepancy between an original image and the resynthesized image with such objects excluded. 
However, utilizing an extra image resynthesis model to detect unexpected objects necessitates a lengthy inference time that is critical in semantic segmentation. 
In the real-world application of semantic segmentation (\textit{e.g.,} autonomous driving), detecting unexpected objects should be finalized in real-time.
Considering such issues, we propose a simple yet effective method 
\textcolor{black}{that can be applied to a given segmentation model without requiring additional training or external datasets.}

\vspace{-0.25cm}
\section{Proposed Method}
\vspace{-0.2cm}
This section presents our approach for detecting unexpected road obstacles.
We first present how we standardize the max logits in Section~\ref{method_SML} and explain how we consider \textcolor{black}{the} local semantics in Section~\ref{method_local_semantic}.

\vspace{-0.2cm}
\subsection{Method Overview}\label{method_overview}
\vspace{-0.2cm}
As our method overview is illustrated in Fig.~\ref{fig:method_overview}, we first obtain the max logits and standardize them, based on the finding that the max logits have their own ranges according to the predicted classes. 
These different ranges cause unexpected objects (pixels in blue boxes) predicted as a certain class to have higher max logit values (\textit{i.e.,} lower anomaly scores) than in-distribution pixels in other classes.
This issue is addressed by standardizing the max logits in a class-wise manner since it enables to reflect the relative meanings within each predicted class.

Then, we remove the false positives (pixels in green boxes) in boundary regions.
Generally, false positives in boundary pixels have lower prediction scores than neighboring in-distribution pixels.
We reduce such false positives by iteratively updating boundary pixels using anomaly scores of neighboring non-boundary pixels. 
Additionally, there exist a non-trivial number of pixels that have significantly different anomaly scores compared to their neighboring pixels, which we term as \emph{irregulars} (pixels in yellow boxes).
Based on the intuition that local consistency (\textit{i.e.,} neighboring pixels sharing similar semantics) exists among pixels in a local region, we apply the smoothing filter with broad receptive fields.
Note that we use \emph{the negative value of the final SML} as the anomaly score.

The following describes the process of how we obtain the max logit and the prediction at each pixel with a given image and the number of pre-defined classes.
Let $\mathbf{X} \in \mathbb{R}^{3\times{H}\times{W}}$ and $C$ denote the input image and the number of pre-defined classes, where $H$ and $W$ are the image height, and width, respectively.
The logit output $\mathbf{F} \in \mathbb{R}^{C\times{H}\times{W}}$ can be obtained from the segmentation network before the softmax layer.
Then, the max logit $\boldsymbol{L} \in \mathbb{R}^{H\times{W}}$ and prediction $\boldsymbol{\hat{Y}} \in \mathbb{R}^{H\times{W}}$ at each location $h$, $w$ are defined as 
\begin{equation}\label{eq_max_logit}
    \vspace*{-0.3cm}
    \boldsymbol{L}_{h,w} = \max_{c} {\mathbf{F}_{c,h,w}}
\end{equation}
\begin{equation}\label{eq_prediction}
    \boldsymbol{\hat{Y}}_{h,w} = \argmax_{c}{\mathbf{F}_{c,h,w}},
\end{equation}
where $c \in \{1, ..., C\}$.

\subsection{Standardized Max Logits (SML)}\label{method_SML}
As described in Fig.~\ref{fig:analysis_box_plot}, standardizing the max logits aligns the distributions of max logits in a class-wise manner.
For the standardization, we obtain the mean $\mu_c$ and variance $\sigma_c^2$ of class $c$ from the training samples.
With the max logit $\boldsymbol{L}_{h,w}$ and the predicted class $\boldsymbol{\hat{Y}}_{h,w}$ from the Eqs.~\eqref{eq_max_logit} and~\eqref{eq_prediction}, we compute the mean $\mu_c$ and variance $\sigma_c^2$ by
\begin{equation}\label{eq_mean}
\vspace*{-0.1cm}
\mu_c = \frac{\sum_i{\sum_{h,w}{\boldsymbol{\mathbbm{1}}(\boldsymbol{\hat{Y}}^{(i)}_{h,w} = c)\cdot{\boldsymbol{L}^{(i)}_{h,w}}}}}
{\sum_i{\sum_{h,w}{\boldsymbol{\mathbbm{1}}(\boldsymbol{\hat{Y}}^{(i)}_{h,w} = c)}}}
\end{equation}
\begin{equation}\label{eq_variance}
\vspace*{-0.1cm}
\sigma_c^2 = \frac{\sum_i{\sum_{h,w}{\boldsymbol{\mathbbm{1}}(\boldsymbol{\hat{Y}}^{(i)}_{h,w} = c)\cdot(\boldsymbol{L}^{(i)}_{h,w} - \mu_c)^2}}}
{\sum_i{\sum_{h,w}{\boldsymbol{\mathbbm{1}}(\boldsymbol{\hat{Y}}^{(i)}_{h,w} = c)}}},
\end{equation}
where $i$ indicates the $i$-th training sample and $\boldsymbol{\mathbbm{1}}( \cdot )$ represents the indicator function.

Next, we standardize the max logits by the obtained statistics.
The SML $\boldsymbol{S} \in \mathbb{R}^{H\times{W}}$ in a test image at each location $h$, $w$ is defined as
\vspace*{-0.1cm}
\begin{equation}\label{eq_sml}
\vspace*{-0.1cm}
\boldsymbol{S}_{h,w} = \frac{\boldsymbol{L}_{h,w} - \mu_{\boldsymbol{\hat{Y}}_{h,w}}}{\sigma_{\boldsymbol{\hat{Y}}_{h,w}}}.
\end{equation}

\subsection{Enhancing with Local Semantics}\label{method_local_semantic}
We explain how we apply iterative boundary suppression and dilated smoothing by utilizing the local semantics.
\begin{table*}[ht!]
\vspace*{-0.1cm}
\begin{center}
\footnotesize
\newcolumntype{a}{>{\columncolor[gray]{0.9}}c}
\begin{tabular}{c|c|c|c|c|a|c|a|c}
\toprule
\multirow{2}{*}{Models} & \multicolumn{2}{c|}{Additional training} & \multirow{2}{*}{\shortstack{Utilizing\\OoD Data}} & \multirow{2}{*}{mIoU} & \multicolumn{2}{c|}{FS Lost \& Found} & \multicolumn{2}{c}{FS Static}  \\
\cline{2-3}\cline{6-9}
                        & Seg. Network           &  Extra Network &                           &                       & \textbf{AP} $\uparrow$& FPR$_{95} \downarrow$                    & \textbf{AP} $\uparrow$& FPR$_{95} \downarrow$ \\
\drule
MSP~\cite{baseline}                                       & \textcolor{red}{\boldxmark}             & \textcolor{red}{\boldxmark}             & \textcolor{red}{\boldxmark}
                                                          & 80.30                                   & 1.77                                    & 44.85          & 12.88          & 39.83 \\ 
                                                          
Entropy~\cite{baseline}                                   & \textcolor{red}{\boldxmark}             & \textcolor{red}{\boldxmark}             & \textcolor{red}{\boldxmark}
                                                          & 80.30                                   & 2.93                                    & 44.83          & 15.41          & 39.75 \\
                                                          
Density - Single-layer NLL~\cite{fishyscapes}             & \textcolor{red}{\boldxmark}             & \textcolor{forestgreen}{\boldcheckmark} & \textcolor{red}{\boldxmark}
                                                          & 80.30                                   & 3.01                                    & 32.90          & 40.86          & 21.29 \\
                                                          
kNN Embedding - density~\cite{fishyscapes}                & \textcolor{red}{\boldxmark}             & \textcolor{red}{\boldxmark}             & \textcolor{red}{\boldxmark}
                                                          & 80.30                                   & 3.55                                    & 30.02          & 44.03          & 20.25\\                                                        
Density - Minimum NLL~\cite{fishyscapes}                  & \textcolor{red}{\boldxmark}             & \textcolor{forestgreen}{\boldcheckmark} & \textcolor{red}{\boldxmark}
                                                          & 80.30                                   & 4.25                                    & 47.15          & 62.14          & 17.43 \\
                                                          
Density - Logistic Regression~\cite{fishyscapes}          & \textcolor{red}{\boldxmark}             & \textcolor{forestgreen}{\boldcheckmark} & \textcolor{forestgreen}{\boldcheckmark}
                                                          & 80.30                                   & 4.65                                    & 24.36          & 57.16          & 13.39 \\
                                                          
Image Resynthesis~\cite{resynthesis}                      & \textcolor{red}{\boldxmark}             & \textcolor{forestgreen}{\boldcheckmark} & \textcolor{red}{\boldxmark}
                                                          & 81.40                                   & 5.70                                    & 48.05          & 29.60          & 27.13 \\

Bayesian Deeplab~\cite{bayesian_deeplab}                  & \textcolor{forestgreen}{\boldcheckmark} & \textcolor{red}{\boldxmark}             & \textcolor{red}{\boldxmark}
                                                          & 73.80                                   & 9.81                                    & 38.46          & 48.70          & 15.50 \\
                                                          
OoD Training - Void Class                                 & \textcolor{forestgreen}{\boldcheckmark} & \textcolor{red}{\boldxmark}             & \textcolor{forestgreen}{\boldcheckmark}
                                                          & 70.40                                   & 10.29                                   & 22.11          & 45.00          & 19.40 \\
                                                          
\textbf{Ours}                                             & \textcolor{red}{\boldxmark}             & \textcolor{red}{\boldxmark}             & \textcolor{red}{\boldxmark}
                                                          & 80.33                                   & \textbf{31.05}                          & \textbf{21.52} & \textbf{53.11} & \textbf{19.64}\\
                                                          
Discriminative Outlier Detection Head~\cite{dense}        & \textcolor{forestgreen}{\boldcheckmark} & \textcolor{forestgreen}{\boldcheckmark} & \textcolor{forestgreen}{\boldcheckmark}
                                                          & 79.57                                   & 31.31                                   & 19.02          & 96.76          & 0.29  \\
                                                          
Dirichlet Deeplab~\cite{prior_network}                    & \textcolor{forestgreen}{\boldcheckmark} & \textcolor{red}{\boldxmark}             & \textcolor{forestgreen}{\boldcheckmark}
                                                          & 70.50                                   & 34.28                                   & 47.43          & 31.3           & 84.60 \\
\bottomrule
\end{tabular}
\end{center}
\vspace*{-0.65cm}
\caption{Comparison with previous approaches reported in Fishyscapes Leaderboard. Models are sorted by the AP scores in Fishyscapes Lost \& Found test set. We achieve a new state-of-the-art performance among the approaches that do not require additional training on the segmentation network or OoD data on Fishyscapes Lost \& Found dataset. Bold fonts indicate the highest performance in its evaluation metric among approaches that do not 1) retrain segmentation networks, 2) train extra networks, and 3) utilize OoD data.}
\label{tab_FS_leaderboard}
\vspace*{-0.7cm}
\end{table*}

\vspace{-0.4cm}
\subsubsection{Iterative boundary suppression}\label{method_boundary}
\begin{figure}[t!]
\begin{center}
    \includegraphics[width=\linewidth]{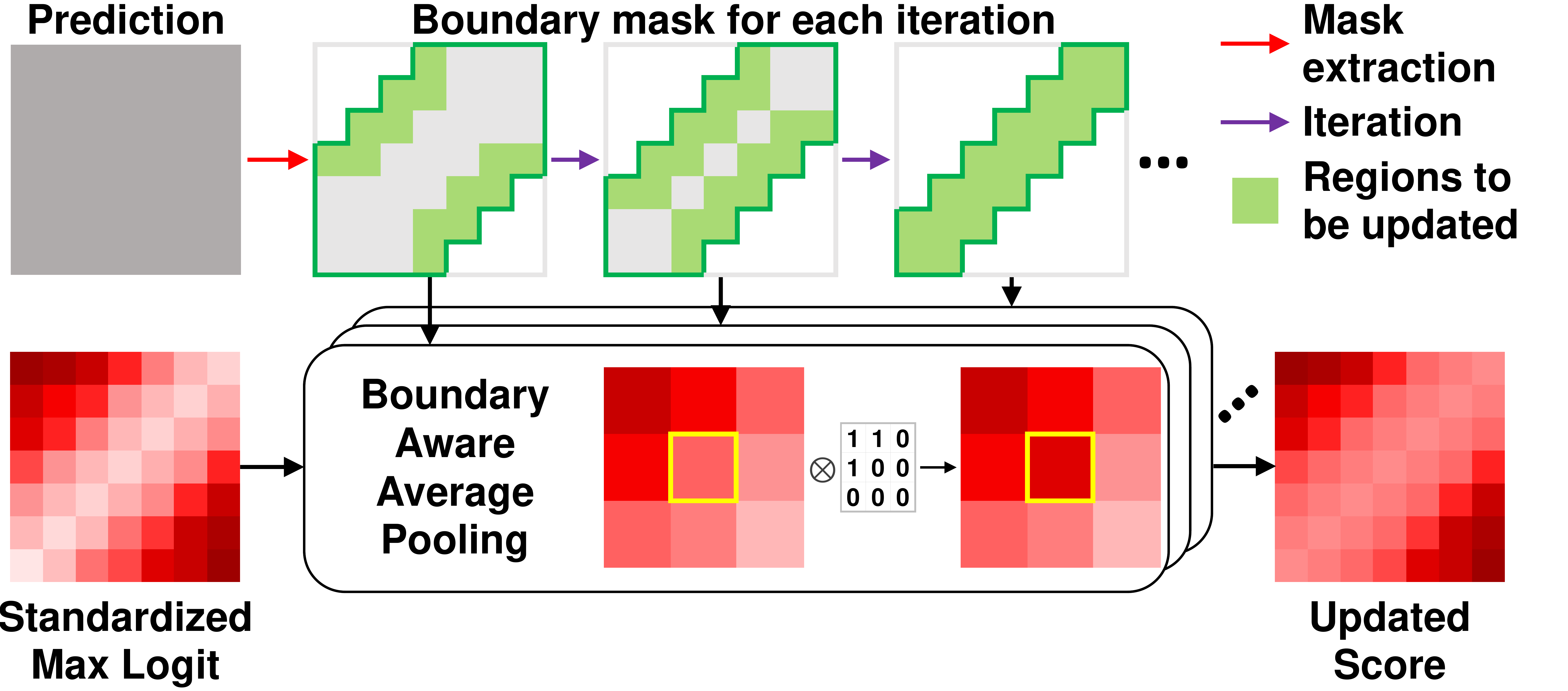}
\end{center}
\vspace*{-0.6cm}
\caption{How iterative boundary suppression works. After standardizing the max logits, we apply average pooling by only using the SMLs of non-boundary pixels (\textit{i.e.,} boundary-aware average pooling) for several iterations. The boundary mask is obtained from a prediction output of a segmentation network.}
\label{fig:method_boundary_suppression}
\vspace*{-0.6cm}
\end{figure}

\vspace{-0.1cm}
To address the problem of wrongly predicting the boundary regions as false positives and false negatives, we iteratively suppress the boundary regions.
Fig.~\ref{fig:method_boundary_suppression} illustrates the process of iterative boundary suppression.
We gradually propagate the SMLs of the neighboring non-boundary pixels to the boundary regions, starting from the outer areas of the boundary (green-colored pixels) to inner areas (gray-colored pixels). To be specific, we assume the boundary width as a particular value and update the boundaries by iteratively reducing the boundary width at each iteration.
This process is defined as follows.
With a given boundary width $r_i$ at the $i$-th iteration and the semantic segmentation output $\boldsymbol{\hat{Y}}$, we obtain the non-boundary mask $\boldsymbol{M}^{(i)} \in \mathbb{R}^{H\times{W}}$ at each pixel $h$, $w$ as
\begin{equation}\label{eq_boundary_mask}
\boldsymbol{M}^{(i)}_{h,w} = \begin{cases}
0, & \text{if $^\exists{h^\prime, w^\prime}\ \  \text{\textit{s.t.,}}\  \boldsymbol{\hat{Y}}_{h, w} \neq \boldsymbol{\hat{Y}}_{h^\prime, w^\prime}$} \\
1, & \text{otherwise}
\end{cases}\quad,
\end{equation}
for $^\forall{h^\prime, w^\prime}$ that satisfies $|h - h^\prime| + |w - w^\prime| \leq r_i$.

Next, we apply the boundary-aware average pooling on the boundary pixels as shown in Fig.~\ref{fig:method_boundary_suppression}.
This applies average pooling on a boundary pixel only with the SMLs of neighboring non-boundary pixels.
With the boundary pixel $b$ and its receptive field $\mathcal{R}$, the boundary-aware average pooling (BAP) is defined as
\begin{equation}\label{eq_boundary_suppression}
BAP(\boldsymbol{S}^{(i)}_\mathcal{R}, \boldsymbol{M}^{(i)}_{\mathcal{R}}) = \frac{\sum_{h,w}{\boldsymbol{S}^{(i)}_{h,w} \times \boldsymbol{M}^{(i)}_{h,w}}}{\sum_{h,w}{\boldsymbol{M}^{(i)}_{h,w}}},
\end{equation}
where $\boldsymbol{S}^{(i)}_\mathcal{R}$ and $\boldsymbol{M}^{(i)}_\mathcal{R}$ denote the patch of receptive field $\mathcal{R}$ on $\boldsymbol{S}^{(i)}$ and $\boldsymbol{M}^{(i)}$, and $(h, w) \in \mathcal{R}$ enumerates the pixels in $\mathcal{R}$.
Then, we replace the original value at the boundary pixel $b$ using the newly obtained one.
We iteratively apply this process for $n$ times by reducing the boundary width by $\Delta{r}=2$ at each iteration. We also set the size of receptive field $\mathcal{R}$ as $3\times3$. In addition, we empirically set the number of iterations $n$ and initial boundary width $r_0$ as 4 and 8.


\vspace{-0.4cm}
\subsubsection{Dilated smoothing}\label{method_smoothing}
\vspace{-0.2cm}
Since iterative boundary suppression only updates boundary pixels, the irregulars in the non-boundary regions are not addressed. 
Hence, we address these pixels by smoothing them using the neighboring pixels based on the intuition that the local consistency exists among the pixels in a local region.
\textcolor{black}{In addition, if the adjacent pixels used for iterative boundary suppression do not have sufficiently low or high anomaly scores, there may still exist boundary pixels that remain as false positives or false negatives even after the process.}
\textcolor{black}{In this regard, we broaden the receptive fields of the smoothing filter using dilation~\cite{dilation} to reflect the anomaly scores beyond boundary regions.}

For the smoothing filter, we leverage the Gaussian kernel since it is widely known that the Gaussian kernel removes noises~\cite{gaussian_blur}.
With a given standard deviation $\sigma$ and convolution filter size $k$, the kernel weight $\boldsymbol{K} \in \mathbb{R}^{k\times{k}}$ at location $i$, $j$ is defined as
\begin{equation}\label{eq_gaussian_kernel}
\boldsymbol{K}_{i,j} = \frac{1}{2\pi\sigma^2}\exp{(-\frac{\Delta{i}^2 + \Delta{j}^2}{2\sigma^2})},
\end{equation}
where $\Delta{i}=i-\frac{(k-1)}{2}$ and $\Delta{j}=j-\frac{(k-1)}{2}$ are the displacements of location $i$, $j$ from the center. In our setting, we set the kernel size $k$ and $\sigma$ to 7 and 1, respectively. Moreover, we empirically set the dilation rate as 6.

\begin{table*}[ht!]
\vspace*{-0.1cm}
\begin{center}
\footnotesize
\begin{tabular}{c|c|c|c|c|c|c|c|c|c|c}
\toprule
\multirow{2}{*}{Models}  & \multirow{2}{*}{mIoU} & \multicolumn{3}{c|}{FS Lost \& Found}   & \multicolumn{3}{c|}{FS Static} & \multicolumn{3}{c}{Road Anomaly}  \\
\cline{3-11}
 & & AUROC $\uparrow$ & AP $\uparrow$ & FPR$_{95}$ $\downarrow$ & AUROC $\uparrow$ & AP $\uparrow$ & FPR$_{95}$ $\downarrow$ & AUROC $\uparrow$ & AP $\uparrow$ & FPR$_{95}$ $\downarrow$ \\
\drule
MSP~\cite{baseline}                             & 80.33                    & 86.99          & 6.02    & 45.63        & 88.94          & 14.24  & 34.10       & 73.76           & 20.59          & 68.44        \\
Max Logit~\cite{maxlogit}                       & 80.33                    & 92.00          & 18.77   & 38.13        & 92.80          & 27.99  & 28.50       & 77.97           & 24.44          & 64.85        \\
Entropy                                         & 80.33                    & 88.32          & 13.91   & 44.85        & 89.99          & 21.78  & 33.74       & 75.12           & 22.38          & 68.15        \\
kNN Embedding - Density~\cite{fishyscapes}      & 80.30                    & -              & 4.1     & 22.30        & -              & -      & -           & -               & -              & -            \\
$^\dagger$SynthCP$^*$~\cite{synthesize_compare} & 80.33                    & 88.34          & 6.54    & 45.95        & 89.90          & 23.22  & 34.02       & 76.08           & 24.86          & 64.69  \\
\textbf{Ours}                & 80.33  & \textbf{96.88} & \textbf{36.55} & \textbf{14.53} & \textbf{96.69} & \textbf{48.67} & \textbf{16.75} & \textbf{81.96} & \textbf{25.82} & \textbf{49.74} \\  
\bottomrule
\end{tabular}
\end{center}
\vspace*{-0.7cm}
\caption{Comparison with other baselines in the Fishyscapes validation sets and the Road Anomaly dataset. $^\dagger$ denotes that the results are obtained from the official code with our pre-trained backbone and $^*$ denotes that the model requires additional learnable parameters. Note that the performance of kNN Embedding - Density is provided from the Fishyscapes~\cite{fishyscapes} team.}
\label{tab_FS_LAF_val}
\vspace*{-0.6cm}
\end{table*}

\vspace{-0.3cm}
\section{Experiments}
\vspace{-0.2cm}
This section describes the datasets, experimental setup, and quantitative and qualitative results. 

\vspace{-0.1cm}
\subsection{Datasets}
\vspace{-0.1cm}

\paragraph{Fishyscapes Lost \& Found~\cite{fishyscapes}} is a high-quality image dataset containing real obstacles on the road. This dataset is based on the original Lost \& Found~\cite{lost_and_found} dataset. 
The original Lost \& Found is collected with the same setup as Cityscapes~\cite{cityscapes}, which is a widely used dataset in urban-scene segmentation. It contains real urban images with 37 types of \textcolor{black}{unexpected} road obstacles and 13 different street scenarios (\textit{e.g.,} different road surface appearances, strong illumination changes, and etc). 
Fishyscapes Lost \& Found further provides the pixel-wise annotations for 1) unexpected objects, 2) objects with pre-defined classes of Cityscapes~\cite{cityscapes}, and 3) void (\textit{i.e.,} objects neither in pre-defined classes nor unexpected objects) regions. 
This dataset includes a public validation set of 100 images and a hidden test set of 275 images for the benchmarking.

\vspace{-0.3cm}
\paragraph{Fishyscapes Static~\cite{fishyscapes}} is constructed based on the validation set of Cityscapes~\cite{cityscapes}. \textcolor{black}{Regarding the objects in the PASCAL VOC~\cite{pascal} as unexpected objects, they are overlaid on the Cityscapes validation images by using various blending techniques to match the characteristics of Cityscapes.}
This dataset contains 30 publicly available validation samples and 1,000 test images that are hidden for benchmarking.

\vspace{-0.4cm}
\paragraph{Road Anomaly~\cite{resynthesis}} contains images of unusual dangers which vehicles confront on roads. 
It consists of 60 web-collected images with anomalous objects (\textit{e.g.,} animals, rocks, and etc.) on roads with a resolution of $1280\times{720}$.
This dataset is challenging since it contains various driving circumstances such as diverse scales of anomalous objects and adverse road conditions.  

\vspace{-0.2cm}
\subsection{Experimental Setup}
\vspace{-0.1cm}
\paragraph{Implementation Details}
We adopt DeepLabv3+~\cite{deepv3+} with ResNet101~\cite{resnet} backbone for our segmentation architecture with the output stride set to 8.  
We train our segmentation networks on Cityscapes~\cite{cityscapes} which is one of the widely used datasets for urban-scene segmentation. 
We use the same pre-trained network for all experiments.

\vspace{-0.4cm}
\paragraph{Evaluation Metrics}
For the quantitative results, we compare the performance by the area under receiver operating characteristics (AUROC) and average precision (AP). 
In addition, we measure the false positive rate at a true positive rate of 95\% (FPR$_{95}$) since the rate of false positives in high-recall areas is crucial for safety-critical applications. \textcolor{black}{For the qualitative analysis, we visualize the prediction results using the threshold at a true positive rate of 95\% (TPR$_{95}$).}

\vspace{-0.4cm}
\paragraph{Baselines}
\textcolor{black}{
We compare ours with the various approaches reported in the Fishyscapes leaderboard.
We also report results on the Fishyscapes validation sets and Road Anomaly with previous approaches that do not utilize external datasets or require additional training for fair comparisons.
Additionally, we compare our method with approaches that are not reported in the Fishyscapes leaderboard. 
Thus, we include the previous method using max logit~\cite{maxlogit} and SynthCP~\cite{synthesize_compare} that leverages an image resynthesis model for such comparison.
Note that SynthCP requires training of additional networks.}

\vspace{-0.2cm}
\subsection{Evaluation Results}
\vspace{-0.2cm}
This section provides the quantitative and qualitative results. We first show the results on Fishyscapes \textcolor{black}{datasets} and Road Anomaly, and then present the comparison results with various backbone networks. 
\textcolor{black}{Additionally, we report the computational cost and the qualitative results by comparing with previous approaches.}

\vspace{-0.4cm}
\subsubsection{Comparison on Fishyscapes Leaderboard}
\vspace{-0.2cm}
Table~\ref{tab_FS_leaderboard} shows \textcolor{black}{the leaderboard result} on the test sets of Fishyscapes Lost \& Found and Fishyscapes Static.
\textcolor{black}{The Fishyscapes Leaderboard categorizes approaches by checking whether they require retraining of segmentation networks or utilize OoD data.
In this work, we add the \emph{Extra Network} column under the \emph{Additional Training} category. 
Extra networks refer to the extra learnable parameters that need to be trained using a particular objective function other than the one for the main segmentation task. 
Utilizing extra networks may require a lengthy inference time, which could be critical for real-time applications such as autonomous driving.
Considering such importance, we add this category for the evaluation.}

As shown in Table~\ref{tab_FS_leaderboard}, we achieve a new state-of-the-art performance on the Fishyscapes Lost \& Found dataset with a large margin, compared to the previous models that do not require additional training \textcolor{black}{of the segmentation network} and external datasets.
Additionally, we even outperform 6 previous approaches in Fishyscapes Lost \& Found and 5 models in Fishyscapes Static which fall into at least one of the two categories.
Moreover, as discussed in the previous work~\cite{fishyscapes}, retraining the segmentation network with additional loss terms impair the original segmentation performance(\textit{i.e.,} mIoU) as can be shown in the cases of Bayesian Deeplab~\cite{bayesian_deeplab}, Dirichlet Deeplab~\cite{prior_network}, and OoD Training with void class in Table~\ref{tab_FS_leaderboard}.
This result is publicly available on the Fishyscapes benchmark website.

\vspace{-0.4cm}
\subsubsection{Comparison on Fishyscapes validation sets and Road Anomaly}
\vspace{-0.2cm}
\textcolor{black}{For a fair comparison, we compare our method on Fishyscapes validation sets and Road Anomaly with previous approaches which do not require additional training and OoD data.}
As shown in Table~\ref{tab_FS_LAF_val}, our method outperforms other previous methods in the three datasets with a large margin.
\textcolor{black}{Additionally, our method achieves a significantly lower FPR$_{95}$ compared to previous approaches.}

\vspace{-0.3cm}
\subsubsection{Qualitative Analysis}
\begin{figure*}[t!]
  \includegraphics[width=\linewidth]{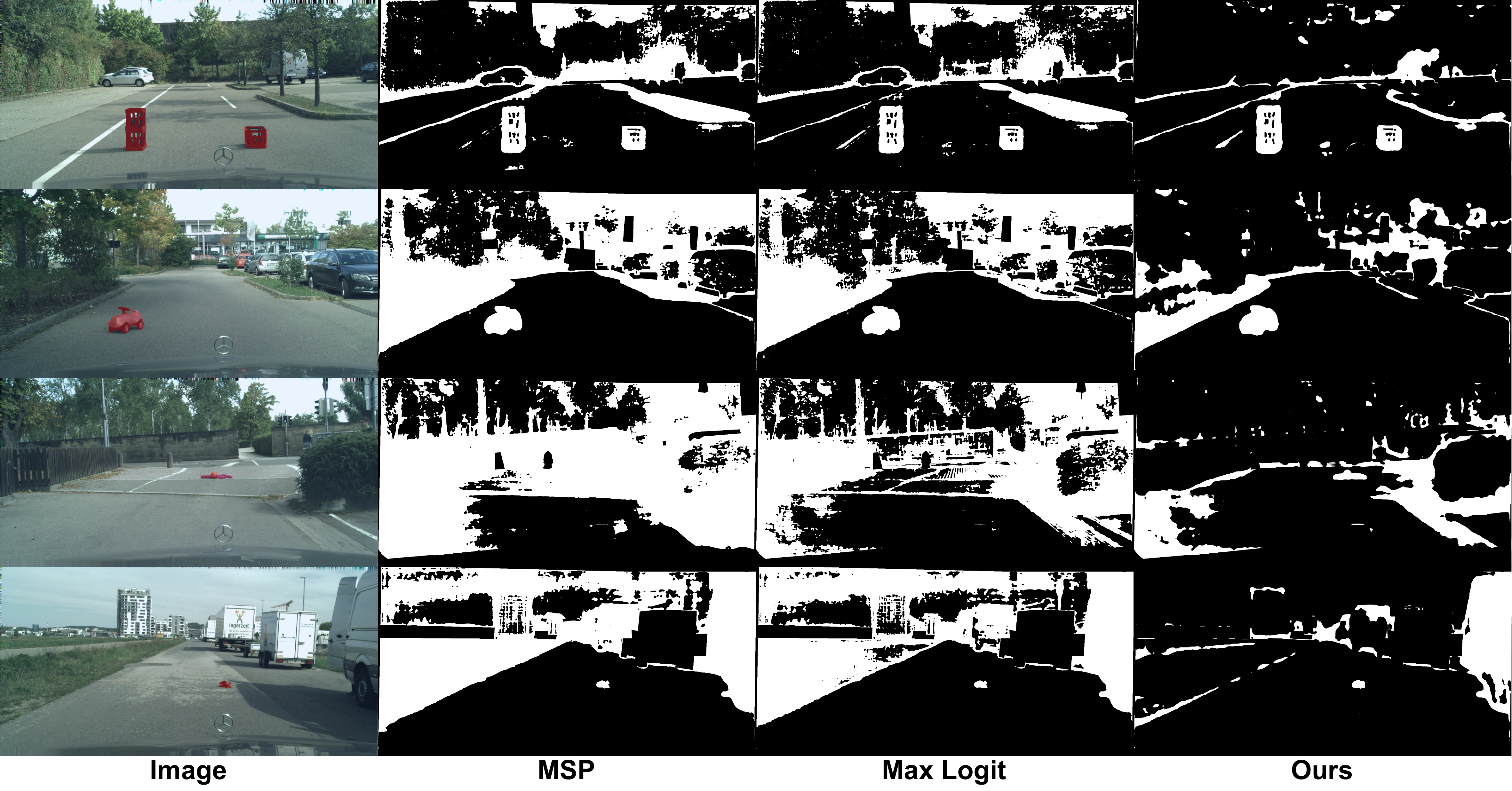}
  \vspace*{-0.75cm}
    \caption{
    Unexpected objects detected with TPR$_{95}$. We compare our method with MSP~\cite{baseline} and max logit~\cite{maxlogit}. White pixels indicate objects which are identified as unexpected objects. Our method significantly reduces the number of false positive pixels compared to the two approaches.
    }
    \vspace*{-0.7cm}
    \label{fig:analysis_tpr}
\end{figure*}

\vspace{-0.2cm}
Fig.~\ref{fig:analysis_tpr} visualizes the pixels detected as unexpected objects (\textit{i.e.,} white \textcolor{black}{regions}) with the TPR at 95\%. 
While previous approaches using MSP~\cite{baseline} and max logit~\cite{maxlogit} require numerous \textcolor{black}{in-distribution pixels} to be detected as unexpected, our method does not.
\textcolor{black}{To be more specific, regions that are less confident (\textit{e.g.,} boundary pixels) are detected as unexpected in MSP~\cite{baseline} and max logit~\cite{maxlogit}.}
However, our method clearly reduces such false positives which can be confirmed by the significantly reduced number of white \textcolor{black}{regions}. 

\vspace{-0.2cm}
\section{\textcolor{black}{Discussion}}
\vspace{-0.2cm}
In this section, we conduct an in-depth analysis on the effects of our proposed method along with the ablation studies.

\begin{table}[h!]
\vspace*{-0.1cm}
\begin{center}
\footnotesize
\begin{tabular}{c|c|c|c}
\toprule
Models  & AUROC $\uparrow$ & AP $\uparrow$ & FPR$_{95}$ $\downarrow$ \\
\drule
Max Logit                         & 92.00           & 18.77          & 38.13 \\
SML                               & 96.54           & 27.61          & 15.46 \\
SML + B Supp.                     & 96.82           & 31.63          & 14.58 \\
SML + D. Smoothing                & 96.70           & 36.00          & 15.65 \\
SML + B Supp. + D. Smoothing      & \textbf{96.89}  & \textbf{36.55} & \textbf{14.53} \\ 
\bottomrule
\end{tabular}
\end{center}
\vspace{-0.6cm}
\caption{Ablation study on our proposed methods. B Supp. and D. Smoothing refer to iterative boundary suppression and dilated smoothing, respectively.}
\label{tab_ablation_ours}
\vspace{-0.4cm}
\end{table}

\vspace{-0.2cm}
\subsection{Ablation Study}
\vspace{-0.1cm}
Table~\ref{tab_ablation_ours} describes the effect of each proposed method in our work with the Fishyscapes Lost \& Found validation set.
SML achieves a significant performance gain over using the max logit~\cite{maxlogit}. 
Performing iterative boundary suppression on SMLs improves the overall performance (\textit{i.e.,} 4\% increase in AP and 1\% decrease in FPR$_{95}$).
On the other hand, despite the increase in AP, performing dilated smoothing on SMLs without iterative boundary suppression results in an unwanted slight increase in FPR$_{95}$. 
The following is the possible reason for the result. When dilated smoothing is applied without iterative boundary suppression, the anomaly scores of non-boundary pixels may be updated with those of boundary pixels. 
Since the non-boundary pixels of in-distribution objects have low anomaly scores compared to the boundaries, it may increase false positives.
Such an issue is addressed by performing iterative boundary suppression before applying dilated smoothing.
After the boundary regions are updated with neighboring non-boundary regions, dilated smoothing increases the overall performance without such error propagation.

\vspace{-0.2cm}
\subsection{Analysis}
\vspace{-0.1cm}
This section provides an in-depth analysis on the effects on segmentation performance, comparison with various backbones, and comparison on computational costs.


\begin{table}[h!]
\vspace{-0.3cm}
\begin{center}
\footnotesize
\begin{tabular}{c|c|c|c|c}
\toprule
Model & Original & MSP & Max Logit & Ours\\
\drule
mIoU (\%)                   & 80.33     & 19.22           & 26.19           & \textbf{68.65} \\ 
\bottomrule
\end{tabular}
\end{center}
\vspace{-0.6cm}
\caption{mIoU on the Cityscapes validation set with the unexpected obstacle detection threshold at TPR$_{95}$ on Fishyscapes Lost \& Found validation set.}
\vspace{-0.5cm}
\label{tab_miou_with_ood}
\end{table}

\vspace{-0.3cm}
\subsubsection{Effects on the segmentation performance}
\vspace{-0.2cm}
Table~\ref{tab_miou_with_ood} shows the mIoU on the Cityscapes validation set with the detection threshold at TPR$_{95}$. By applying the detection threshold, the segmentation model predicts a non-trivial amount of in-distribution pixels as the unexpected ones. Due to such false positives, the mIoU of all methods decreased from the original mIoU of 80.33\%.
To be more specific, using MSP~\cite{baseline} and max logit~\cite{maxlogit} result in significant performance degradation. On the other hand, our approach maintains a reasonable performance of mIoU even with outstanding unexpected obstacle detection performance. This table again demonstrates the practicality of our work since it both shows reasonable performance in the segmentation task and the unexpected obstacle detection task.

\begin{table}[h!]
\vspace*{-0.3cm}
\begin{center}
\footnotesize
\begin{tabular}{c|c|c|c|c|c}
\toprule
Backbone & Models & mIoU & AUROC $\uparrow$ & AP $\uparrow$ & FPR$_{95}$ $\downarrow$ \\
\drule
\multirow{3}{*}{\shortstack{MobileNet\\V2~\cite{mobile}}}
& MSP           & \multirow{3}{*}{75.70} & 86.00          & 2.60    & 48.05 \\
& Max Logit     &                        & 91.89          & 7.15    & 36.24 \\
& \textbf{Ours} &                        & \textbf{96.18} & \textbf{16.95} & \textbf{16.63} \\
\midrule
\multirow{3}{*}{\shortstack{ShuffleNet\\V2~\cite{shuffle}}}
& MSP           & \multirow{3}{*}{72.71} & 86.33          & 4.06    & 45.68 \\
& Max Logit     &                        & 90.06          & 8.67    & 45.36 \\
& \textbf{Ours} &                        & \textbf{95.26} & \textbf{14.42} & \textbf{23.17} \\
\midrule
\multirow{3}{*}{\shortstack{ResNet50\\~\cite{resnet}}}
& MSP           & \multirow{3}{*}{77.76} & 86.25          & 3.50    & 45.03 \\
& Max Logit     &                        & 89.47          & 8.95    & 48.99 \\
& \textbf{Ours} &                        & \textbf{95.24} & \textbf{18.54} & \textbf{19.57} \\
\bottomrule
\end{tabular}
\end{center}
\vspace{-0.6cm}
\caption{Comparison with MSP and max logit on Fishyscapes Lost \& Found dataset. The backbone networks are trained with the output stride of 16.}
\label{tab_FS_backbone}
\vspace{-0.55cm}
\end{table}

\vspace{-0.4cm}
\subsubsection{Comparison with various backbones}
\vspace{-0.2cm}
\textcolor{black}{Since our method does not require additional training or extra OoD datasets, our method can be adopted and used easily on any existing pre-trained segmentation networks.}
\textcolor{black}{To verify the wide applicability of our approach, we report the performance of identifying anomalous objects with various backbone networks including MobileNetV2~\cite{mobile}, ShuffleNetV2~\cite{shuffle}, and ResNet50~\cite{shuffle}}.
\textcolor{black}{As shown in Table~\ref{tab_FS_backbone}, our method significantly outperforms the other approaches~\cite{baseline, maxlogit} using the same backbone network with a large improvement in AP. 
This result clearly demonstrates that our method is applicable widely regardless of the backbone network.}

\begin{table}[h!]
\vspace{-0.3cm}
\begin{center}
\footnotesize
\begin{tabular}{c|c|c}
\toprule
Models & GFLOPs & Infer. Time (ms) \\
\drule
ResNet-101~\cite{resnet}                   & 2139.86 & 60.54  \\
Ours (SML)                                 & 2139.86 & 61.41  \\
Ours (SML + B. Supp .)                     & 2140.01 & 74.66  \\ 
Ours (SML + B. Supp. + D. Smoothing)       & 2140.12 & 75.02  \\ 
SynthCP~\cite{synthesize_compare}          & 4551.11 & 146.90 \\
\bottomrule
\end{tabular}
\end{center}
\vspace{-0.6cm}
\caption{Comparison of computational cost. Metrics are measured with the image size of $2048\times{1024}$ on NVIDIA GeForce RTX 3090 GPU. The inference time is averaged over 100 trials.}
\label{tab_inference_time}
\vspace{-0.6cm}
\end{table}

\vspace{-0.35cm}
\subsubsection{Comparison on computational cost}
\vspace{-0.2cm}
To demonstrate that our method requires a negligible amount of computation cost, we report GFLOPs (\textit{i.e.,} the number of floating-point operations used for computation) and the inference time.
\textcolor{black}{As shown in Table~\ref{tab_inference_time}, our method requires only a minimal amount of computation cost regarding both GFLOPs and the inference time compared to the original segmentation network, ResNet-101~\cite{resnet}.
Also, among several studies which utilize additional networks, we compare with a recently proposed approach~\cite{synthesize_compare} that leverages an image resynthesis model.
Our approach requires substantially less amount of computation cost compared to SynthCP~\cite{synthesize_compare}.}

\begin{table}[h!]
\vspace{-0.3cm}
\begin{center}
\footnotesize
\begin{tabular}{c|c|c|c}
\toprule
Models  & $\Delta$AUROC $\uparrow$ & $\Delta$AP $\uparrow$ & $\Delta$FPR$_{95}$ $\downarrow$ \\
\drule
MSP + B. Supp. + D. S.                   & -0.60           & 1.08           & 3.24 \\ 
Max Logit + B. Supp. + D. S.             & -0.51          & -1.45           & 2.60 \\
SML + B. Supp. + D. S.                   & \textbf{0.35}  & \textbf{8.95} & \textbf{-0.93} \\ 
\bottomrule
\end{tabular}
\end{center}
\vspace{-0.6cm}
\caption{Comparison of metric gains after iterative boundary suppression and dilated smoothing on MSP, max logit, and SML. B Supp. and D. S refer to iterative boundary suppression and dilated smoothing, respectively.}
\label{tab_ablation_post_processing}
\vspace{-0.3cm}
\end{table}

\vspace{-0.2cm}
\subsection{Effects of Standardized Max Logit}
\vspace{-0.1cm}
Table~\ref{tab_ablation_post_processing} describes how SML enables applying iterative boundary suppression and dilated smoothing.
Applying iterative boundary suppression and dilated smoothing on other approaches does not improve the performance or even aggravates in the cases of MSP~\cite{baseline} and max logit~\cite{maxlogit}.
On the other hand, it significantly enhances the performance when applied to SML. 
The following are the possible reasons for such observation.
As aforementioned, the overconfidence of the softmax layer elevates the MSPs of anomalous objects.
Since the MSPs of anomalous objects and in-distribution objects are not distinguishable enough, applying iterative boundary suppression and dilated smoothing may not improve the performance.

Additionally, iterative boundary suppression and dilated smoothing require the values to be scaled since it performs certain computations with the values.
In the case of using max logits, the values of each predicted class differ according to the predicted class.
Performing the iterative boundary suppression and dilated smoothing in such a case aggravates the performance because the same max logit values in different classes represent different meanings according to their predicted class.
SML aligns the differently formed distributions of max logits which enables to utilize the values of neighboring pixels with certain computations. 

\vspace{-0.3cm}
\section{Conclusions}
\vspace{-0.2cm}
In this work, we proposed a~\textit{simple yet effective} method for identifying unexpected obstacles on roads that do not require external datasets or additional training. 
Since max logits have their own ranges in each predicted class, we aligned them via standardization, which improves the performance of detecting anomalous objects. 
Additionally, based on the intuition that pixels in a local region share local semantics, we iteratively suppressed the boundary regions and removed irregular pixels that have distinct values compared to neighboring pixels via dilated smoothing. 
With such a straightforward approach, we achieved a new state-of-the-art performance on Fishyscapes Lost \& Found benchmark.
Additionally, extensive experiments with diverse datasets demonstrate the superiority of our method to other previous approaches.
Through the visualizations and in-depth analysis, we verified our intuition and rationale that standardizing max logit and considering the local semantics of neighboring pixels indeed enhance the performance of identifying unexpected obstacles on roads.
However, there still remains room for improvements; 1) dilated smoothing might remove unexpected obstacles that are as small as noises, and 2) the performance depends on the distribution of max logits obtained from the main segmentation networks.

We hope our work inspires the following researchers to investigate such practical methods for identifying anomalous objects in urban-scene segmentation which is crucial in safety-critical applications. 

\vspace{-0.3cm}
\section{Acknowledgement}
\vspace{-0.2cm}
We deeply appreciate Hermann Blum and FishyScapes team for their sincere help in providing the baseline performances and helping our team to update our model on the FishyScapes Leaderboard.

This work was supported by the Institute of Information \& communications Technology Planning \& Evaluation (IITP) grant funded by the Korean government(MSIT) (No. 2019-0-00075, Artificial Intelligence Graduate School Program(KAIST) and No. 2020-0-00368, A Neural-Symbolic Model for Knowledge Acquisition and Inference Techniques) and the National Research Foundation of Korea (NRF) grant funded by the Korean government (MSIT) (No. NRF-2019R1A2C4070420).

\newpage
{\small
\bibliographystyle{ieee_fullname}
\bibliography{egbib}
}

\def\thesection{\Alph{section}}

\setcounter{section}{0}
\section{Supplementary Material}
This supplementary presents the quantitative results on different architectures, hyper-parameter impacts, implementation details, and qualitative results.

\subsection{Effects on Different Architecture and Backbone}
This section presents the quantitative results of different architecture and backbone (\textit{i.e.,} EfficientPS~\cite{efficientps} and ResNeSt~\cite{resnest}) on the FishyScapes Lost \& Found validation set.
As shown in Table~\ref{supp_tab_different_architectures}, our approach outperforms all other methods in both cases. However, the amount of performance increase is not strictly correlated with the downstream task performance, as also pointed out in the previous work~\cite{downstream_task_correlation1, downstream_task_correlation2}.

\begin{table}[h!]
\begin{center}
\footnotesize
\begin{tabular}{c|c|c|c|c|c}
\toprule
Architectures & mIoU & Methods & AUROC $\uparrow$ & AP $\uparrow$ & FPR$_{95}$ $\downarrow$ \\
\drule

\multirow{3}{*}{\shortstack{$^\dagger$EfficientPS\\~\cite{efficientps}}} & \multirow{3}{*}{79.3} & MSP       & 84.41 & 1.46 & 61.03 \\
                             &                                                    & Max Logit & 89.39 & 3.83 & 48.75 \\
                             &                                                    & \textbf{Ours}      & \textbf{94.17} & \textbf{5.93} & \textbf{21.93}\\
\midrule
\multirow{3}{*}{\shortstack{DeeplabV3+\\ w/ ResNeSt\\~\cite{resnest}}} & \multirow{3}{*}{79.1} & MSP       & 87.23 & 7.89  & 57.67 \\
                                                      &                                      & Max Logit & 91.91 & 22.58 & 51.12 \\
                                                      &                                      & \textbf{Ours}      & \textbf{95.32} & \textbf{31.38} & \textbf{30.37} \\
\bottomrule
\end{tabular}
\end{center}
\vspace*{-0.5cm}
\caption{Results of EfficientPS and DeeplabV3+ with ResNeSt backbone on Fishyscapes Lost \& Found validation set. $^\dagger$ denotes the results are obtained from the official code with their pre-trained networks.}
\label{supp_tab_different_architectures}
\vspace*{-0.5cm}
\end{table}

\subsection{Analysis on Hyper-parameters} \label{supp_hyper_parameters}
This section analyzes the impact of hyper-parameters in our proposed method through ablation studies on FishyScapes Lost\&Found validation set.

\paragraph{Number of iterations $\boldsymbol{n}$}
We report the quantitative results according to the number of iterations $n$ used in iterative boundary suppression, described in Section \textcolor{red}{3.3.1} of the main paper. 
Note that we set the initial boundary width $r_0$ to $2n$ so that $\Delta{r}=\lfloor\frac{r_0}{n}\rfloor$ equals 2 since we intend to reduce the width by 1 from each side of the boundary.
As shown in Table~\ref{supp_tab_boundary_dilation_ablation}, the performances in all metrics consistently increase as $n$ increases up to $n=4$. 
While AUROC and FPR$_{95}$ are improved at $n=5$, AP rather aggravates.
Since the number of in-distribution and unexpected pixels are unbalanced, we choose AP for our primary metric, which is invariant to the data imbalance, as done in Fishyscapes. Hence, we use $n=4$ in our work.

\begin{table}[h!]
\begin{center}
\footnotesize
\begin{tabular}{c|c|c|c}
\toprule
Iterations &  AUROC $\uparrow$ & AP $\uparrow$ & FPR$_{95}$ $\downarrow$ \\
\drule
$n=1$ & 96.73 & 36.26 & 15.48 \\
$n=2$ & 96.78 & 36.44 & 15.19 \\
$n=3$ & 96.84 & 36.54 & 14.86 \\
$\boldsymbol{n=4}$ & 96.89 & \textbf{36.55} & 14.53 \\
$n=5$ & \textbf{96.93} & 36.44 & \textbf{14.22} \\
\bottomrule
\end{tabular}
\end{center}
\vspace*{-0.5cm}
\caption{Quantitative results with respect to $n$ on Fishyscapes Lost \& Found. Results are obtained after standardizing the max logit, iterative boundary suppression, and dilated smoothing.}
\label{supp_tab_boundary_dilation_ablation}
\vspace*{-0.5cm}
\end{table}




\paragraph{Dilation rate $\boldsymbol{d}$}
We present the quantitative results with respect to the dilation rate $d$ used in dilated smoothing, described in Section \textcolor{red}{3.3.2} of the main paper.
As shown in Table~\ref{supp_tab_dilation_ablation}, taking wider receptive fields improves the performance in AP up to $d=6$. 
However, if the size of the receptive field increases further (\textit{e.g.,} after $d=7$), the performance rather degrades, indicating that a proper size of a receptive field is crucial in properly capturing the consistent local patterns. 

\begin{table}[h!]
\vspace*{-0.1cm}
\begin{center}
\footnotesize
\begin{tabular}{c|c|c|c}
\toprule
Dilation &  AUROC $\uparrow$ & AP $\uparrow$ & FPR$_{95}$ $\downarrow$ \\
\drule
$d=1$ & 96.86 & 33.25 & 14.50 \\
$d=2$ & 96.90 & 34.61 & 14.36 \\
$d=3$ & 96.92 & 35.57 & \textbf{14.33} \\
$d=4$ & \textbf{96.93} & 36.15 & 14.39 \\
$d=5$ & 96.92 & 36.46 & 14.47 \\
$\boldsymbol{d=6}$ & 96.89 & \textbf{36.55} & 14.53 \\
$d=7$ & 96.86 & 36.47 & 14.57 \\
$d=8$ & 96.81 & 36.28 & 14.66 \\
$d=9$ & 96.76 & 35.99 & 14.91 \\
$d=10$& 96.70 & 35.64 & 15.31 \\
\bottomrule
\end{tabular}
\end{center}
\vspace*{-0.2cm}
\caption{Quantitative results according to the dilation rate $d$ on Fishyscapes Lost \& Found. Results are obtained after standardizing the max logit, iterative boundary suppression, and dilated smoothing.}
\label{supp_tab_dilation_ablation}
\vspace*{-0.2cm}
\end{table}

\subsection{Further Implementation Details}
We adopt DeepLabV3+~\cite{deepv3+} as our segmentation network architecture and mainly use ResNet101~\cite{resnet} as the backbone for most of the experiments. 
Note that, as already shown in the main paper, our proposed method is model-agnostic and achieves the best performance with the MobileNetV2~\cite{mobile}, ShuffleNetV2~\cite{shuffle}, and ResNet50~\cite{resnet} backbones compared to MSP~\cite{baseline} and max logit~\cite{maxlogit}.

The model is trained with an output stride of 8 and the batch size of 8 for 60,000 iterations with an initial learning rate of 1e-2 and momentum of 0.9. In addition, we apply the polynomial learning rate scheduling~\cite{polynomial_scheduling} with the power of 0.9 and the standard cross-entropy loss with the auxiliary loss proposed in PSPNet~\cite{pspnet}, where the auxiliary loss weight $\lambda$ is set to 0.4. Moreover, in order to prevent the model from overfitting, we apply color and positional augmentations such as color jittering, Gaussian blur, random scaling with the range of [0.5, 2.0], random horizontal flipping, and random cropping. We adopt class-uniform sampling~\cite{class_uniform_and_urban_scene, class_uniform_2} with a rate 0.5.

As aforementioned, we set the number of boundary iterations $n$, the initial boundary width $r_0$, and the dilation rate $d$ as 4, 8, and 6, respectively. 
Additionally, we set the sizes of the boundary-aware average pooling kernel and the smoothing kernel size as $3\times3$ and $7\times7$, respectively.

\subsection{Qualitative Results}
This section presents the additional qualitative results. 
We first demonstrate the qualitative results of our methods and then their comparisons with other baselines.
We use the threshold at $TPR_{95}$ and visualize the predicted in-distribution and unexpected pixels as black and white, respectively.

\begin{figure*}[t!]
  \includegraphics[width=\linewidth]{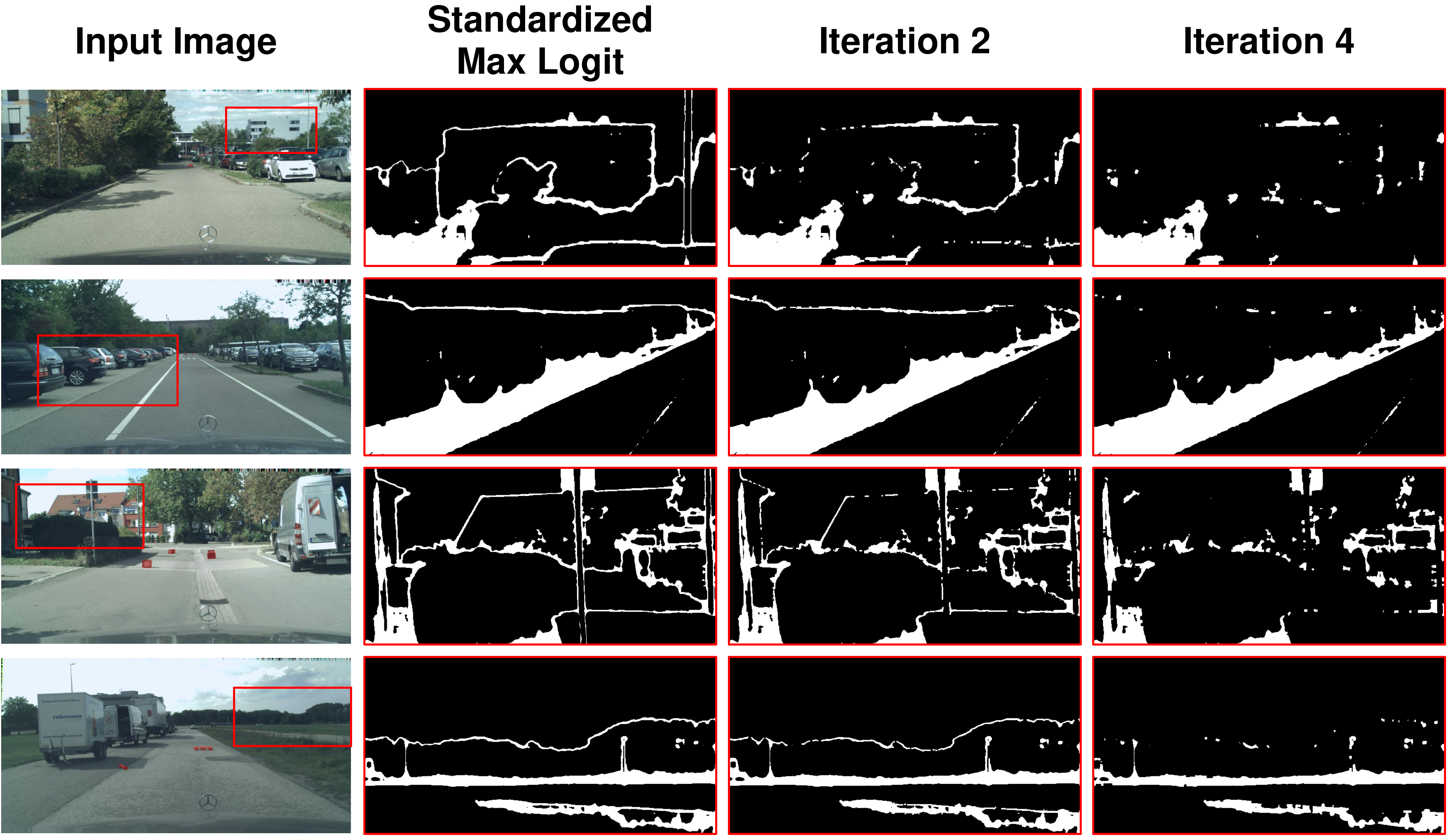}
    \caption{Qualitative results of applying standardized max logit and iterative boundary suppression with iteration 2 and 4, respectively. We report the images of Fishyscapes Lost \& Found. The white pixels indicate the pixels predicted as unexpected.}
    \label{fig:analysis_false_positive_boundary}
\end{figure*}

\vspace{-0.4cm}
\paragraph{Our results}
Fig.~\ref{fig:analysis_false_positive_boundary} presents the qualitative results of applying iterative boundary suppression to show the effectiveness of removing the false positives (\textit{i.e.,} in-distribution pixels detected as unexpected). 
We zoom in particular regions with the red boxes to show the changes in detail.
After applying iterative boundary suppression, we significantly remove the false positives in boundary regions. 

\begin{figure*}[t!]
  \includegraphics[width=\linewidth]{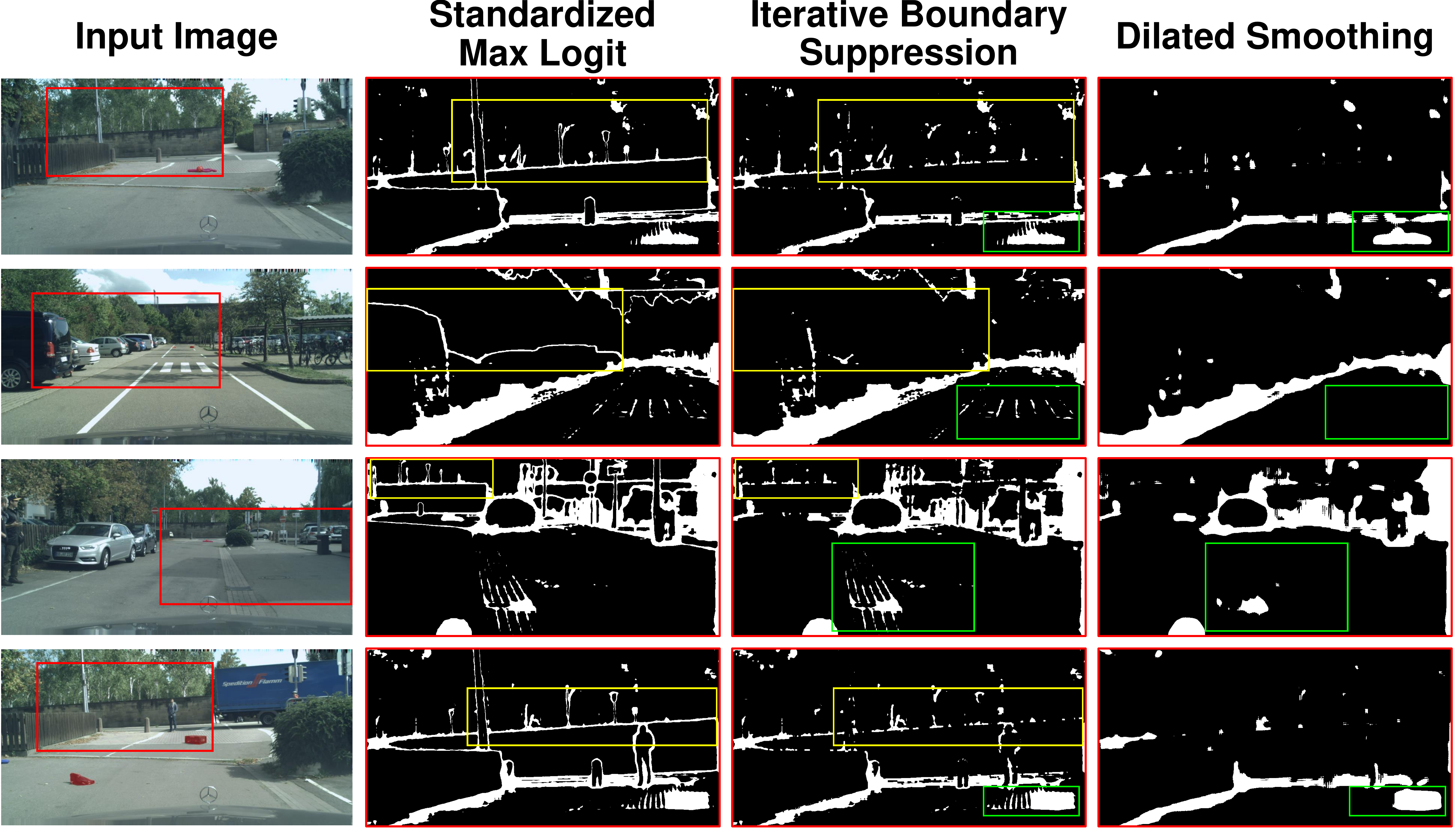}
    \caption{Qualitative results of applying standardized max logit, iterative boundary suppression, and dilated smoothing, respectively. We report the images of Fishyscapes Lost \& Found. Yellow boxes and green boxes show that the false positives are effectively removed by applying iterative boundary suppression and dilated smoothing, respectively. The white pixels indicate the pixels predicted as unexpected.}
    \label{fig:analysis_false_positive_total}
    \vspace{0.5cm}
\end{figure*}

Additionally, Fig.~\ref{fig:analysis_false_positive_total} describes the results of applying all of our methods. 
The false positives in the boundary regions (\textit{e.g.,} white pixels in the yellow boxes) are removed after applying iterative boundary suppression.
Also, as shown in the green boxes, applying dilated smoothing effectively removes the false positives in the non-boundary regions.

\begin{figure*}[t!]
  \includegraphics[width=\linewidth]{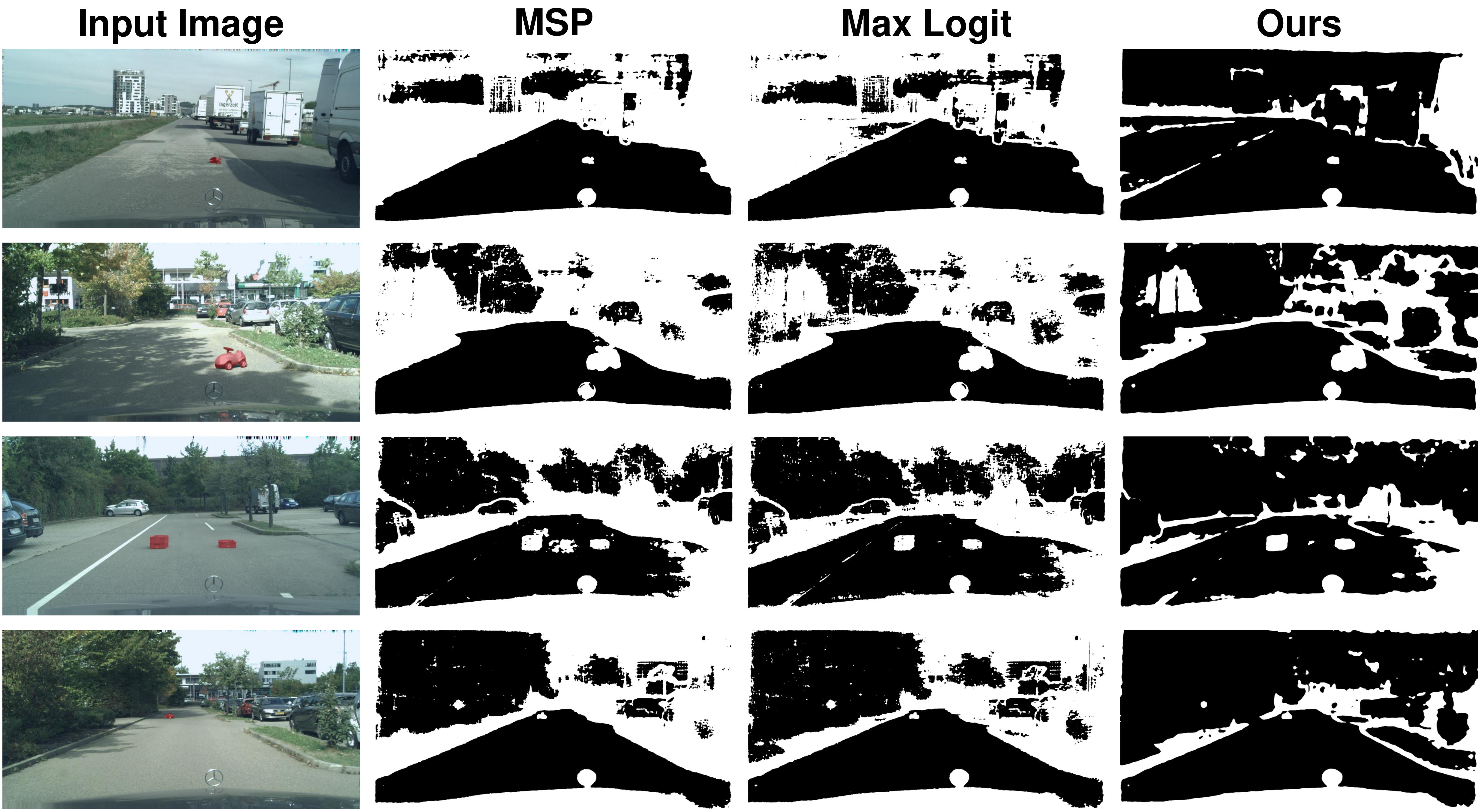}
    \caption{Comparison with MSP, max logit, and ours on Fishyscapes Lost \& Found dataset. The white pixels indicate the pixels predicted as unexpected.}
    \label{fig:analysis_false_positive_lf}
\end{figure*}

\begin{figure*}[t!]
  \includegraphics[width=\linewidth]{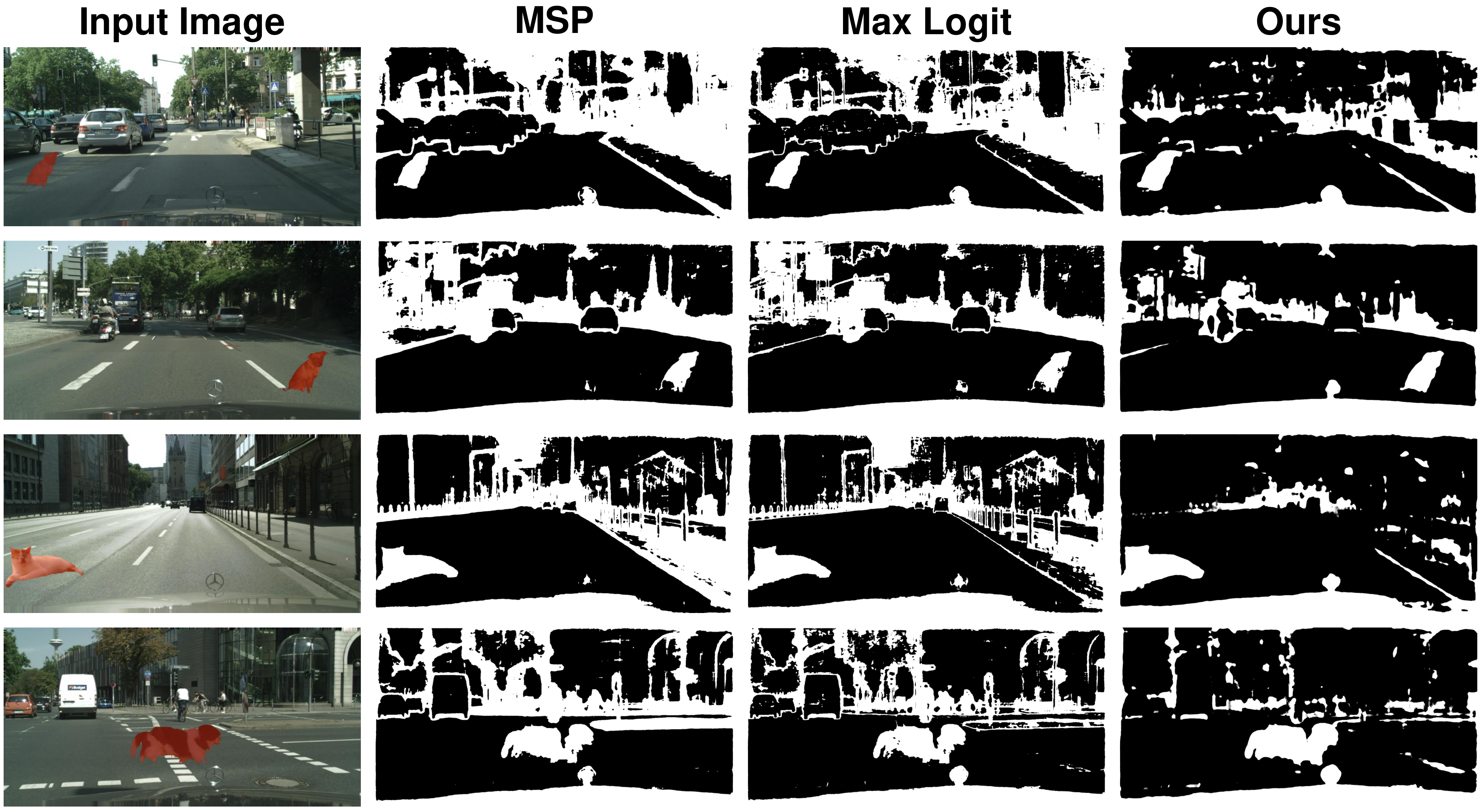}
    \caption{Comparison with MSP, max logit, and ours on Fishyscapes Static dataset. The white pixels indicate the pixels predicted as unexpected.}
    \label{fig:analysis_false_positive_static}
\end{figure*}

\vspace{-0.4cm}
\paragraph{Comparison with other approaches}
We compare our method with MSP~\cite{baseline} and max logit~\cite{maxlogit} by showing qualitative results.
Figs.~\ref{fig:analysis_false_positive_lf} and \ref{fig:analysis_false_positive_static} show the results obtained from Fishyscapes Lost \& Found and Fishsyscapes Static, respectively.
Since we visualize the images with the threshold at $TPR_{95}$, most of the pixels in unexpected objects are identified.
However, using MSP and max logit generate a substantial amount of false positives.
In contrast, our method produces a negligible amount of false positives, which demonstrates our effectiveness.

\vspace*{-0.5cm}

\end{document}